\pdfoutput=1

\documentclass[11pt]{article}

\usepackage[]{ACL2023}

\usepackage{times}
\usepackage{latexsym}

\usepackage[T1]{fontenc}

\usepackage[utf8]{inputenc}

\usepackage{microtype}

\usepackage{inconsolata}

\usepackage{amsmath}
\usepackage{amssymb}
\usepackage{booktabs}
\usepackage{multirow}
\usepackage{graphicx}
\usepackage[ruled,vlined]{algorithm2e}
\usepackage{algorithmic}
\usepackage{color}

\title{Enhancing Personalized Dialogue Generation with Contrastive Latent Variables: Combining Sparse and Dense Persona}

\author{Yihong Tang$^{1}$, Bo Wang$^{2,}$\thanks{$^*$Corresponding author.} , Miao Fang$^4$, \\ {\bf Dongming Zhao$^3$, Kun Huang$^3$, Ruifang He$^2$, Yuexian Hou$^2$} \\
        $^1$School of New Media and Communication, Tianjin University, Tianjin, China \\ 
        $^2$College of Intelligence and Computing, Tianjin University, Tianjin, China \\ 
        $^3$AI Lab, China Mobile Communication Group Tianjin Co., Ltd. \\
        $^4$School of Computer and Communication Engineering, \\Northeastern University at Qinhuangdao, Qinghuangdao, China \\
        \texttt{\{toyhom, bo\_wang\}@tju.edu.cn}
}

\begin{document}
\maketitle

\begin{abstract}
The personalized dialogue explores the consistent relationship between dialogue generation and personality. Existing personalized dialogue agents model persona profiles from three resources: sparse or dense persona descriptions and dialogue histories. However, sparse structured persona attributes are explicit but uninformative, dense persona texts contain rich persona descriptions with much noise, and dialogue history query is both noisy and uninformative for persona modeling. In this work, we combine the advantages of the three resources to obtain a richer and more accurate persona. We design a \textbf{C}ontrastive \textbf{L}atent \textbf{V}ariable-based model (\textbf{CLV}) that clusters the dense persona descriptions into sparse categories, which are combined with the history query to generate personalized responses. Experimental results on Chinese and English datasets demonstrate our model's superiority in personalization.
\end{abstract}

\section{Introduction}

In order to develop personalized dialogue agents, current approaches enhance the personality of generated responses mainly utilizing three kinds of resources: (1) Defined sparse persona attributes ~\citep{zhang-2018-personalizing,song2019exploiting,wolf-2019-TransferTransfo,liu-2020-impress,song-2021-bob}; (2) Dense persona description texts ~\citep{Qian-2018-assign,Zheng-2020-Pre-Training,song-2021-bob}; (3) Historical queries of current dialogue ~\citep{li-2016-persona,ma-2021-onechatbot}. Each of the three resources has its advantages and disadvantages. 

Sparse persona attributes (e.g., gender, age) are highly interpretable and have high information utilization, but the information is limited and cannot express complex persona features. Dense persona description text contains rich and flexible persona information but suffers from noisy expressions. Modeling personality directly from dialogue histories is free of additional persona information, but the persona information in history queries is both noisy and uninformative.

To address these issues, in this paper, we improve personalized dialogue generation by combining the advantages of the three resources. We design a contrastive latent variable (CLV)-based model that clusters the dense persona descriptions into sparse categories, which are combined with the history query to generate personalized responses. Specifically, first, the dialog's latest query and response together with dense persona description texts are encoded. Then the recognition distribution of query and response is jointly modeled with a pre-designed dual conditional variational autoencoder (CVAE~\citep{sohn-2015-learning}). Simultaneously, the persona information is automatically separated into multiple parts to participate in the above process in parallel. These partitioned persona pieces of information are considered to hide different angles of portrayal. This process is also reinforced by contrastive learning. Next, a decider decides which category of persona information is used for persona modeling. Finally, a personalized generator combines the history query and additional persona information for response generation. Without explicit supervised signals, we design a pseudo-labeling and joint training method to train the decider.

Our contributions are summarized as follows:


(1) We design a framework named CLV based on contrastive latent variables to combine the advantages of three persona resources for personalized dialogue generation. The framework contains a self-separation algorithm and a decider, which are jointly trained to work in conjunction with each other. In this way, our work can both extract information more efficiently from the cluttered persona description text and not require persona information in the inference phase. 


(2) Under the designed CLV-based framework, we propose a self-separation algorithm to mine and categorize dense persona description text into sparse persona profiles. Furthermore, a decider is proposed to decide whether the dialogue should involve persona information and choose appropriate persona profiles among the persona profiles generated by the self-separation algorithm. This process helps to improve the consistency of personalized dialogue generation.

(3) We conduct extensive experiments on the Chinese and English personalized dialogue datasets to demonstrate our model's superiority. We also propose a refined evaluation framework for personalized dialogue generation, which considers the consistency, coherence, and diversity of dialogue generation at the same time.

\begin{figure*}[htbp]
\centering
\includegraphics[height=.3\textheight,width=.9\textwidth]{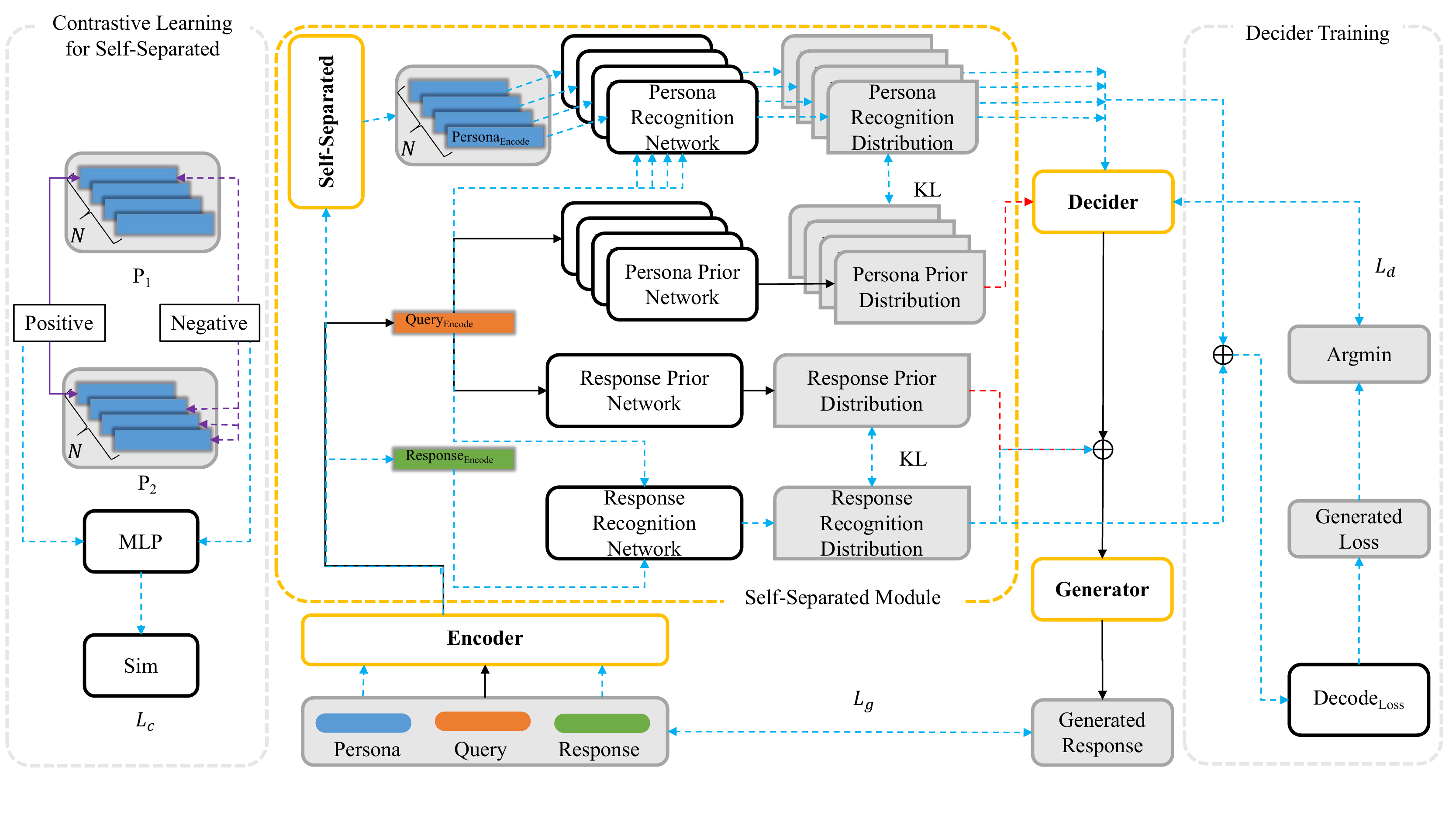}
\caption{The overview structure of the proposed model. Connections with dashed blue lines only appear during training, connections with dashed red lines only appear during inference, and connections with solid black lines indicate that they appear during both training and inference phases. The purple lines represent positive and negative example constructions in contrastive learning.}
\label{fig:overview}
\end{figure*}

\section{Related Work}

\paragraph{Personalized Dialogue Generation}

Open-domain dialogue has been studied in depth for a long time~\citep{koehn-2003-statistical, Ni-2021-RecentAI}, and under the influence of the psychological theory, personality has been incorporated into the requirements for dialogue generation. Personalized dialogue generation has three typical approaches: (1) Using well-defined sparse persona attributes (e.g., gender, age), the model can utilize different attributes efficiently and interpretably, and knowledge-enhanced dialogue generation approaches can be borrowed ~\citep{zhang-2018-personalizing,song2019exploiting,wolf-2019-TransferTransfo,liu-2020-impress,bao-2020-plato,song-2021-bob}. However, sparse attributes can only provide little persona information without complex semantics. (2) Mining information from dense textual persona descriptions, which contain rich and deep persona information but are very noisy ~\citep{Qian-2018-assign,song-2020-profile,Zheng-2020-Pre-Training,song-2021-bob}. (3) Implicitly modeling persona profiles from historical dialogue query ~\citep{li-2016-persona,ma-2021-onechatbot,zhong-etal-2022-less}. This approach does not rely on additional persona information, but it is difficult to acquire personality implicitly from dialogue history without reference objects.

\paragraph{Dialogue generation based on CVAE}
Besides personalization, another essential goal of personalized dialogue generation is the diversity of dialog expression. To this end, existing works have explored hidden variable models that model the variables in the dialogue process as Gaussian distributions, which can enhance the diversity of dialogue generation by introducing randomness~\citep{zhao-2017-learning,song2019exploiting,hu-2022-fuse}. In this direction, one typical approach is to include persona information as a condition in regular Seq2Seq constructs and to model responses and queries as recognition distributions in CVAE~\cite{li-2018-generating-classical}; another approach is to combine persona information or other external conditions and responses as generation targets before modeling joint distributions together with queries~\citep{lee-2021-dlvgen}. In addition, many CVAE text generation models focus on other tasks, and they modify model details as well as probability maps for different tasks, which are not considered in this paper.

\section{Methodology}

\subsection{Overview}
Given multi-turn dialogue of two users $u_i, u_j$. The dialogue context of $u_i$ is $U^i = \{(Q_1^i,R_1^i),\cdots,(Q_t^i,R_t^i)\}$. $Q^i$ is the \textit{query} initiated by $u_j$ to $u_i$. The goal of the personalized dialogue is to generate a personalized \textit{response} $R_i$ using the corresponding \textit{personal} information $P_i$ in text form.


The overview of our model is shown in Figure~\ref{fig:overview}. The overall model is composed of four modules: encoder, self-separation module, decider, and generator (marked in Figure~\ref{fig:overview} with orange borders). Specifically, the encoder module encodes dialogue queries, persona information, and responses respectively. The self-separation module separates the persona information in the  hidden sentence vector space to form the grouping of persona information with implicit categories. We use multiple CVAEs to process the grouping persona information and get the grouping latent variables. The decider then automatically selects the latent variable to use from the group and feeds it into the generator along with the query. Finally, the generator autoregressively generates personalized responses based on the query and latent variables.

\subsection{Encoder}\label{sec:encoder}

we use a pre-trained GPT-2 ~\citep{radford2019language} to encode the personal information text $P_i$, dialog query $Q_i$, and dialog response $R_i$. We take the hidden vector of the last time step in the last layer of GPT-2 as the representation of the whole paragraph:
\begin{align}
{p_i} &= \operatorname{GPT-2_{Hidden}}(P_i),\\
{q_i} &= \operatorname{GPT-2_{Hidden}}(Q_i),\\
{r_i} &= \operatorname{GPT-2_{Hidden}}(R_i),
\end{align}
where $p_i, q_i, r_i\in \mathbb{R}^d$, and $d$ is the dimension of the hidden state.

\renewcommand{\algorithmicrequire}{\textbf{Input:}}
\renewcommand{\algorithmicensure}{\textbf{Output:}}
\begin{algorithm}[h]

    \caption{Persona Self-Separation}\label{alg:ss}
    \begin{algorithmic}[1]
    \REQUIRE
    $p \in \mathbb{R}^{1\times d}$ : the vector representation of original sentence;\\
    $N$ : hyper-parameter, the self-separation coefficient;\\
    $d$ : the dimension of the hidden state;
    \ENSURE
    $P_{g} \in \mathbb{R}^{N\times d}$ : vector representations of persona information after processing, in this context, it is the form of a set;\\
        \STATE Initialize $P_{g}$;
        \STATE Set $s \leftarrow $ the integer of $d/N$;
        \FOR {$i=1$ to $N$}
        \STATE Initialize augment vector $c_i\leftarrow (0,0,\ldots,0)_{1\times d}$;
        \STATE Set $c_i\left[(i-1)\times s+1:i\times s\right]\leftarrow (1,1,\ldots,1)_{1\times s}$;
        \STATE $P_{g}\left[i,:\right]\leftarrow \operatorname{MLP}(p+c_i;c_i)$;
        \ENDFOR
    \RETURN $P_{g}$
    \end{algorithmic}
\end{algorithm}

\subsection{Self-Separated Module}
After obtaining the hidden state representation of $P$, $Q$ and $R$, their representation vectors are further processed. As mentioned above, sparse personal information is more explicit and interpretable, while dense information text contains rich information but needs to be more organized. Therefore, referring to the research of ~\citet{sun2021generating}, we propose a self-separation method of persona information, which implicitly divides dense text persona information into $N$ categories:
\begin{align}
P_g &= \operatorname{P-Sepa}(p),
\end{align}
where $P_g = \{p_1,p_2,\cdots,p_N\}$, and $P_g$ represents the persona information after grouping, which is composed of multiple parallel persona information. For the algorithm of P-Sepa, see Algorithm~\ref{alg:ss}.

In order to let the model automatically classify the grouped persona information, we use contrastive learning on the data in the same batch to let the model learn the similarities between the grouped persona information. Specifically, for two data points, $P_g^i$ and $P_g^j$, we use a contrastive loss to help the model better represent group persona information.Following simcse, we denote $h^i_k = f_\theta(p^i_k)$ where $p^i_k \in P_g^i$. Then we get the training objective:
\begin{align}
L_c &= -log\frac{e^{sim(h_k^i,h_k^j)/\tau}}{\sum_{n=1}^{N}e^{sim(h_k^i,h_n^j)/\tau}}, 
\end{align}
where $\tau$ is a temperature hyperparameter and $sim(h_k^i,h_k^j)$ is the cosine similarity. 


The model samples the persona latent variable $z_p$ from the persona distribution and the response latent variable $z_r$ from the potential response distribution. Since $z_p$ and $z_r$ respectively represent different aspects of the generated responses ($z_p$ contains the persona, and $z_r$ captures the specific query-response association), we assume that $z_p$ and $z_r$ are independent of each other, namely $z_p \perp z_r$. So, the response generation process can be said to use the following conditional distribution $p(r,z_p,z_r|q) = p(r|q,z_p,z_r)p (z_p|q)p(z_r|q)$. Our goal is to use the deep learning method to approximate $p(r|q,z_p,z_r)$, $p(z_p|q)$ and $p(z_r|q)$, in which, according to~\citet{zhao-2017-learning} and~\citet{song2019exploiting}, we refer to $p(r|q,z_p,z_r)$ as a response generator and $p_\theta(z_p|q)$, $p_\theta(z_r|q)$ as a \textit{prior network}. In order to approximate the posterior distribution of the true, we refer to $q_\varphi(z_p|q,p)$ and $q_\varphi(z_r|q,r)$ as recognition networks.

We train this CVAE using Stochastic Gradient Variational Bayes(SGVB)~\citep{kingma2013auto} by maximizing the \textit{variational lower bound} of conditional log-likelihood.
Following~\citet{zhao-2017-learning} and~\citet{song2019exploiting}, we assume that potential variables $z_p$ and $z_r$ follows a multivariate Gaussian distribution with the diagonal covariance matrix. The lower bound of the variation of CLV-CVAE can be written as:
\begin{equation}\label{formula:kl}
\small
\begin{aligned}
L_{g} &= E_{q_{\varphi_r}(z_r|q,r);q_{\varphi_r}(z_p|q,p)}(\log p(r|q, z)) \\ &- KL(p_{\theta_q} (z_p|q)||q_{\varphi_r} (z_p|q,p)) \\
&- KL(p_{\theta_r} (z_r|q)||q_{\varphi_r} (z_r|q,r)),
\end{aligned}
\end{equation}

Because we assume that the underlying variables $z_p$ and $z_r$ follow isotropic multivariate gaussian distribution, both recognition networks $q_{\varphi_p}(z_p|q,p) \sim  \mathcal{N}(\mu_p,\sigma_p^2\mathbf{I})$ and $q_{\varphi_r}(z_r|q,r) \sim  \mathcal{N}(\mu_r,\sigma_r^2\mathbf{I})$, both prior networks $p_{\theta_p}(z_p|q) \sim  \mathcal{N}(\mu_p^\prime,\sigma_p^{\prime2}\mathbf{I})$ and $p_{\theta_r}(z_r|q)  \sim  \mathcal{N}(\mu_r^\prime,\sigma_r^{\prime2}\mathbf{I})$. In order to sample $z_p$ and $z_r$ from the prior network and recognition network in training and to make the sampling operation differentiable, using the \textit{reparameterization} technique~\citep{kingma2013auto}, we have:
\begin{align}
\begin{bmatrix}
 \mu_p \\ \sigma_p^2
\end{bmatrix} &= W^{recog}_q
\begin{bmatrix}
 q\\p
\end{bmatrix}+b^{recog}_q, 
\\
\begin{bmatrix}
 \mu_r \\ \sigma_r^2
\end{bmatrix} &= W^{recog}_r
\begin{bmatrix}
 q\\r
\end{bmatrix}+b^{recog}_r, 
\\
\begin{bmatrix}
 \mu_p^\prime \\ \sigma_p^{\prime2}
\end{bmatrix} &= W^{prior}_qq
+b^{prior}_q,  
\end{align}

\begin{align}
\begin{bmatrix}
 \mu_r^\prime \\ \sigma_r^{\prime2}
\end{bmatrix} &= W^{prior}_rr
+b^{prior}_r,
\end{align}
where $p,r,q$ are the representation vectors obtained in Section~\ref{sec:encoder}.

Finally, $z$ is fed into the generator to generate $r$ together with the dialogue query $q$, where: $z=z_p+z_r$.
 How to get the final $z_p$ is explained in detail in Section~\ref{sec:decider}.

\subsection{Decider}\label{sec:decider}

In fact, in order to make the model can find the appropriate persona information, we do not let CLV choose from the grouped persona information directly, but first, use the recognition network or prior network to obtain the grouped persona information latent variables $Z_p^g=\{ z_p^1,z_p^2,\cdots,z_p^N\}$, which is obtained by sampling a set of distributions constructed separately for each vector in $P_g$. Then, the \textit{Decider} is trained to choose between them. We call it the Decider because it also includes the decision not to use personal information.

Specifically, the decider is a classification neural network composed of multi-layer sensing units which use a soft decision method to make a selection. The decider-matrix is composed of classification probability, and the classification probability is multiplied by the grouping persona information latent variable to get the final persona information latent variable $z_p$. For grouped persona information latent variable $Z_p^g$:
\begin{align}
    W_{d} &= \operatorname{Softmax}(\operatorname{MLP}([Z_p^g;q])), \\
    z_p &= W_{d} \cdot Z_p^g,
\end{align}
where $Z_p^g\in \mathbb{R}^{N\times d}$, $W_{d}\in \mathbb{R}^{1 \times N}$ and $z_p\in \mathbb{R}^{d}$.

It is difficult to directly let the decider learn how to choose from the latent variables of grouping persona information generated by sampling the persona distribution of implicit clustering. Therefore, we introduce the pseudo-label to guide the learning of the decider. The more intuitive idea is that if a latent variable in the group of persona information latent variables can achieve a minor decoding loss in the generator, then it may be a better latent variable. Based on this idea, we designed the decision loss to train the decider:
\begin{align}
y &=\operatorname{Argmin}(\operatorname{GPT-2_{Loss}}(Z_p^g)), \label{formula:y}\\
L_d &= -y\log(W_{d}), \label{formula:ld}
\end{align}
where $y$ is the index corresponding to $z_p$ input into the generator to obtain the minimum decoding loss.

\subsection{Generator}
We use a pre-trained GPT-2 as the generator, which uses the dialogue query as input and adds cross-attention to the latent variable $z$:
\begin{align}
    \hat{R} &= \operatorname{GPT-2_{Generator}}(Pre(z),q),
\end{align}
where $Pre(z)$ is the pre-cross attention object added before the standard GPT-2, which autoregressively generates a personalized response $\hat{R}$.

\begin{table}[t]
\centering
\scriptsize
\begin{tabular}{llll}
\toprule
\multicolumn{1}{c}{\textbf{Dataset}} &
  \multicolumn{1}{c}{\textbf{\# Train}} &
  \multicolumn{1}{c}{\textbf{\# Valid}} &
  \multicolumn{1}{c}{\textbf{\# Test}} \\ \midrule
ConvAI2     & 43,410           & 4,213            & 2,138            \\
Baidu PersonaChat    & 376,016         & 19,923 &  4,456    \\ \bottomrule
\end{tabular}
\caption{Statistics of persona dialogue datasets.}
\label{tab:datasets}
\end{table}

\subsection{Training and Optimizer}
In our practice, we find that there are some challenges in training the decider, which is probably the reason for the mutual influence between loss functions. Firstly, there will be conflicts between the KL divergence and the decoding loss of the generator. Secondly, the loss of the decider depends on the dummy label monitoring signal set by us. Finally, for the purpose of implicit clustering of persona information, the contrastive enhancement loss is largely independent of the above losses.

In order to promote gradient learning involving the above loss functions, a joint training process is designed to train CVAE and decider alternately. Specifically, in each training iteration, we first sample query $Q$, response $R$, and persona information $P$ of two data points from batch data $D$, conduct contrastive training on encoders encoding persona information according to the self-separation algorithm~\ref{alg:ss}, and then generate latent variables after self-separation respectively according to the method described in Section ~\ref{sec:decider}. The generator's loss value creates a dummy label y (Eq.~\ref{formula:y}), which is used to train the decider by optimizing the loss $L_d$(Eq.~\ref{formula:ld}).

Further, we traverse $D$, generate a personalized response $R$, and update the generator and CVAE MLP by optimizing loss $L_{g}$ (Eq.~\ref{formula:kl}).

\begin{table*}[t!]
\centering
\small
\resizebox{\textwidth}{!}{
\begin{tabular}{llrrrrrrrr}
\toprule
&  & \multicolumn{3}{c}{Coherence} & \multicolumn{4}{c}{Diversity} & {Consistency} \\  
\cmidrule(lr){3-5}\cmidrule(lr){6-9}\cmidrule(lr){10-10}  
& & {BLEU-1} & {ROUGE-L} & {Coh.Score} &  {C-Dist-1} & {C-Dist-2} & {S-Dist-1} &  {S-Dist-2} &  {Coh-Con.Score} \\ 
\midrule
\multirow{7}[0]{*}{ConvAI2} & Seq2Seq & 3.45$^\dagger$ & 5.45$^\dagger$ & 34.85$^\dagger$ & 1.23$^\dagger$ & 3.84$^\dagger$ & 34.21$^\dagger$ & 61.59$^\dagger$ & 10.85$^\dagger$ \\
 & GPT-2 & 6.77$^\dagger$ & 10.96$^\dagger$ & 56.71$^\dagger$  & 7.35$^\dagger$ & 28.13$^\dagger$ &68.22$^\dagger$ & 88.81$^\dagger$ & 13.29$^\dagger$ \\
 & PerCVAE & 6.89$^\dagger$ & 10.54$^\dagger$& 53.26$^\dagger$ & \textbf{12.57}$^\dagger$ & \textbf{39.54}$^\dagger$ & 67.48$^\dagger$ & 89.46$^\dagger$ & 12.95$^\dagger$ \\
 & BoB & 7.85$^\dagger$ & 12.46$^\dagger$ & 62.47$^\dagger$ & 7.24$^\dagger$ & 26.41$^\dagger$ & 63.85$^\dagger$ & 85.02$^\dagger$ & 15.97$^\dagger$ \\
 & DHAP & 7.21$^\dagger$ & 9.90$^\dagger$ & 64.27$^\dagger$ & 9.24$^\dagger$ & 30.98$^\dagger$ & {69.86}$^\dagger$ & {90.23}$^\dagger$ & 16.04$^\dagger$ \\
 & MSP & {8.19}$^\dagger$ & {11.67}$^\dagger$& {65.81}$^\dagger$ & {10.49}$^\dagger$ & {29.96}$^\dagger$  & {65.79}$^\dagger$ & {89.43}$^\dagger$ & {15.45}$^\dagger$ \\
 & {CLV} (Ours) & \textbf{11.85} & \textbf{15.10}& \textbf{71.72} & {5.63}  & {26.91} & \textbf{71.24} & \textbf{92.89} &  \textbf{23.01} \\ 
\midrule
\multirow{7}[0]{*}{Baidu PersonaChat} & Seq2Seq & 7.14$^\dagger$ & 8.66$^\dagger$ & 40.39$^\dagger$  & 0.97$^\dagger$ & 5.19$^\dagger$ & 29.61$^\dagger$ & 76.65$^\dagger$ &  8.96$^\dagger$ \\
 & GPT-2 & 10.53$^\dagger$ & 11.29$^\dagger$ & 49.37$^\dagger$ & 5.64$^\dagger$ & 24.98$^\dagger$ & 51.93$^\dagger$  & 84.06$^\dagger$ & 12.14$^\dagger$ \\
 & PerCVAE & 10.86$^\dagger$ & 10.44$^\dagger$ & 51.19$^\dagger$ & \textbf{10.39}$^\dagger$ & 27.86$^\dagger$  & 58.24$^\dagger$ & 87.37$^\dagger$ & 11.33$^\dagger$ \\
 & BoB & 14.26$^\dagger$ & 13.30$^\dagger$ & 58.13$^\dagger$ & 5.36$^\dagger$ & 27.45$^\dagger$ & 52.91$^\dagger$ & 82.93$^\dagger$ & 16.33$^\dagger$ \\
 & DHAP & 12.96$^\dagger$ & 12.54$^\dagger$ & 55.21$^\dagger$ & 6.23$^\dagger$ & 25.37$^\dagger$ & 57.09$^\dagger$  & 85.44$^\dagger$ & 12.30$^\dagger$ \\
 & MSP & 15.84$^\dagger$ & 14.06$^\dagger$ & \textbf{61.52}$^\dagger$ & 5.37$^\dagger$ & \textbf{28.41}$^\dagger$ & 54.06$^\dagger$  & 86.24$^\dagger$ & 14.37$^\dagger$ \\
 & {CLV} (Ours) & \textbf{24.77} & \textbf{22.33} & 60.74 & 2.42 & 22.96 & \textbf{60.27} & \textbf{88.15} & \textbf{18.15} \\ 
\bottomrule
\end{tabular}}
\caption{Automatic evaluation on two datasets. The best results are in \textbf{bold}. ``$\dagger$'' indicates that our model passed the t-test with $p$-value $<$ 0.05.}\label{tab:auto}
\end{table*}

\section{Experiments}

\subsection{Datasets}
\textbf{ConvAI2}~\citep{dinan2019second} is an English dataset containing rich personal information, and the dialogues in this dataset are based on the personal facts corresponding to the characters. It is derived from PersonaChat~\citep{zhang-etal-2018-personalizing} and obtained after filtering and refinement. It is a crowdsourced dataset covering rich persona features, and we have processed it to remove some noise.

\textbf{Baidu PersonaChat\footnote{\url{https://www.luge.ai/\#/luge/dataDetail?id=38}}}, which is a personalization dataset collected and open-sourced by Baidu, is similar to ConvAI2, although it's Chinese.

We summarize the key statistics of the two personalized dialogue datasets in Table~\ref{tab:datasets}.
As mentioned earlier, we only use the persona information of the two datasets during training.

\subsection{Baselines}
We compare the proposed model with 6 baselines, which can be classified into 3 categories.

\paragraph{Non-Personalized Approaches} 
\textbf{Seq2Seq} with Attention ~\citep{Sutskever-2014-sequence} is a sequence-to-sequence model with an attention mechanism (Luong et al., 2015). The pre-trained \textbf{GPT-2} ~\citep{radford2019language} performs well in various text generation tasks and is used as a dialogue generation model after training on a dialogue corpus.

\paragraph{Approaches based on Dense Persona Information} 
These methods use persona information to construct knowledge enhancement models, and for better model comparison, we tested these methods using the dialogue history as an approximation of the persona information. \textbf{PerCVAE} ~\citep{zhao-2017-learning} encodes the persona information text as a conditional representation and uses CVAE to generate personalized responses. \textbf{BoB}~\citep{song-2021-bob} uses the Bert model for personalized dialogue generation and integrates the consistency generation task with the consistency inference task jointly to provide insight into the evaluation mechanism of personalized dialogue generation.

\paragraph{The Dialogue History-based Approach} 
\textbf{DHAP} ~\citep{ma-2021-onechatbot} uses historical memory to store and construct dynamic query-aware user profiles from dialogue histories and then uses a personalized decoder to generate responses. \textbf{MSP} ~\citep{zhong-etal-2022-less} enhances personalized dialogue generation by retrieving similar conversations from similar users via User Refiner and Topic Refiner and uses a Token Refiner to find the relevant tokens to be used during training, which is the best overall performance model for persona-free information personalized dialogue generation.

\paragraph{Implementation Details}
are in Appendix~\ref{sec:dps}.

\subsection{Evaluations}

In order to obtain accurate performance comparisons, we use both automatic and human evaluations.

\paragraph{Automatic Evaluation} 
We divide the automatic evaluation methods into three categories in order to evaluate and model the diversity, consistency, and coherence of the generated dialogues.

(1) \textbf{Diversity} Distinct-1/2~\citep{li-2016-diversity} considers the number of single or double frames in the generated responses and is usually used to evaluate diversity. Most experiments do not specify the object of evaluation for Distinct-1/2, whether it is the whole corpus or multiple sentences, so we propose C-Dist-1/2(Corpus-Distinct-1/2) and S-Dist-1/2(Sentence-Distinct-1/2) according to the different objects of evaluation, the former evaluating the dialogue responses generated by the model on the whole test set, and the latter evaluating multiple responses (set to generate five responses in this paper). S-Dist-1/2 provides a better evaluation of whether the model can generate interesting responses in the same situation.

(2) \textbf{Consistency} The personalized dialogue generation task requires consistency between the generated responses and the persona information, and we propose Con.Score (Consistency Score) based on C.score~\citep{madotto-2019-personalizing}, which is obtained based on the referee model and can be defined as:

\begin{equation}
\small
\label{formula:cscore}
\begin{aligned}
\text{Con.Score}(P,Q,R) = \left\{
\begin{aligned}
1 & , & \text{if}\ \text{NLI}(P,Q,R)\ =\ 1\ or\ 2, \\
0 & , & \text{if}\ \text{NLI}(P,Q,R)\ =\ 0.
\end{aligned}
\right.
\end{aligned}
\end{equation}

where the NLI model is a triple classification model and can be found in Appendix~\ref{sec:appendix}.

(3) \textbf{Coherence} BLEU-1~\citep{papineni-2002-bleu} and ROUGE-L~\citep{lin-2004-autorouge} are classical words overlap-based metrics for measuring the similarity between generated responses and factual responses, which we believe can indirectly measure the coherence of dialogues. The reason we didn't look at BLEU-2/3/4 because we think that too much rigid coverage doesn't reflect the coherence of the model. And similar to the Con.Score, we propose the Coh-Con.Score (Coherence-Consistency Score), which is also obtained based on the NLI model:
\begin{equation}
\small
\label{formula:cscore}
\begin{aligned}
\text{Coh-Con.Score}(P,Q,R) = \left\{
\begin{aligned}
0 & , & \text{if}\ \text{NLI}(P,Q,R)\ =\ 0, \\
1 & , & \text{if}\ \text{NLI}(P,Q,R)\ =\ 2.
\end{aligned}
\right.
\end{aligned}
\end{equation}

\paragraph{Human Evaluation} 
 Taking into account the uncertainty of the criteria when evaluating, we perform human evaluations of all models, and we convert the scoring method to a ranking method. Specifically, we extract 100 data points(queries, responses, and persona information) and hire three well-educated annotators to score the responses generated by the different models in a ranking style and to normalize them into specific scores on a scale of $[0, 1]$ at the end. We focus on four aspects: readability, diversity, consistency, and coherence, and ask the evaluators to rank eight options for the seven model-generated responses and the factual responses.

\subsection{Experimental Results}

\paragraph{Automatic Evaluation} 
Table~\ref{tab:auto} shows the performance of all models on different automatic metrics for both Chinese and English datasets, and it can be clearly observed that our CLV model improves on key metrics and these improvements are statistically significant (t-test with $p$-value $<$ 0.05). Specifically, we can observe that: (1) \textbf{Diversity}. CLV shows different results on the two diversity evaluation dimensions. For S-Dist-1/2, CLV leads the other models, which indicates that our model is able to make more diverse and flexible responses compared to other models when facing the same situation. However, C-Dist-1/2 is lower than most models, which indicates that our model makes some sacrifices to improve consistency and coherence, and we will analyze this reason further in Section~\ref{sec:fa}. (2) \textbf{Consistency}. The lead of the consistency personalization metric Con.Score implies that our approach can integrate persona information into the generation, especially when this integration is done without information generation, which is more indicative of the superiority of CLV. (3) \textbf{Coherence}. The performance of our model in coherence is also outstanding, whether it is the coverage index BLEU-1, Rouge-L, or the learning index Coh-Con.Score, which also shows that it is feasible to use the coverage index as a kind of evaluation basis for dialogue coherence. Our task diversity, coherence, and consistency can be used as three key bases for evaluating personalized dialogue generation, and the findings in the experiments suggest that our model is able to produce more personalized responses than all baselines.

\begin{table}[t!]
\centering
\small
\resizebox{\columnwidth}{!}{
\begin{tabular}{lrrrr}
\toprule
Model & Readability & Diversity & Consistency & Coherence \\ \midrule
Seq2Seq & 0.57$^\dagger$ & 0.69$^\dagger$ & 0.11$^\dagger$ & 0.34$^\dagger$ \\
GPT-2  & {0.73}$^\dagger$ &  {0.72}$^\dagger$  & 0.43$^\dagger$ & 0.69$^\dagger$ \\
PerCVAE  & 0.71$^\dagger$ &  {0.82}$^\dagger$  & 0.41$^\dagger$ & 0.65$^\dagger$ \\
BoB & 0.72$^\dagger$ &  0.80$^\dagger$  & 0.57$^\dagger$ & 0.73$^\dagger$ \\
DHAP & 0.77$^\dagger$ &  0.85  & 0.49$^\dagger$ & 0.69$^\dagger$ \\
MSP & 0.75$^\dagger$ &  0.83$^\dagger$  &  {0.51}$^\dagger$ & 0.72$^\dagger$ \\ 
{CLV (N=4)}  &  \textbf{0.79} & \textbf{0.85}  & \textbf{0.61} & \textbf{0.81} \\ \midrule
Ground-Truth &  0.80 &  0.91  &  0.86 & 0.97 \\ \bottomrule
\end{tabular}}
\caption{The result of human evaluation on ConvAI2 dataset. ``$\dagger$'' indicates that our model passed the t-test with $p$-value $<$ 0.05.}
\label{tab:human}
\end{table}

\paragraph{Human Evaluation} 
Human evaluation results on ConvAI2 are shown in Table~\ref{tab:human}. We calculated the Fleiss Kappa among the three annotators and obtained a Kappa of 0.67, which implies that the three annotators are in \textit{substantial agreement}~\citep{landis1977measurement}. In general, the results of human annotations are consistent with the results of automatic evaluations. They both demonstrate the advantages of our model in terms of personalized dialogue generation and basic readability.


\begin{table*}[t]
\centering
\scriptsize
\begin{tabular}{p{.11\linewidth}p{.22\linewidth}p{.22\linewidth}p{.22\linewidth}}
\toprule
Persona  & I have a happy family. I sleep about six hours a day. I usually like to play the guitar. I work in a factory. I like listening to \textcolor{red}{music}. & I am a \textcolor{red}{school teacher}. I love to taste good food. I teach grade five in the \textcolor{red}{Affiliated Primary School of Renmin University}. & I'm afraid of heights. I don't believe in religion. I'm a \textcolor{red}{lawyer}. \\ \midrule
Query  & Yes, and what are your hobbies? & So you're from Beijing. I'm from Beijing, too. & A case? Are you a policeman? \\ \midrule Golden-Response & My life is relatively simple, in addition to making music, to help the factory. What do you like to do in your spare time? & Really? Are you from Beijing, too? & No, I'm a lawyer. I specialize in property disputes.  \\ \midrule
GPT-2 & I like running and my body feels good. & Nice. I own a bakery. & No, I'm not. I'm a fireman.  \\ \midrule
MSP & I usually like to dance, I will dance with other anime anime anime anime voices. & Haha, I went to school in \textcolor{red}{Beijing}. I want to be a doctor in the future. & No, just talking and playing.  \\ 
\midrule
BoB & I love studying and I like \textcolor{red}{music} very much. & No, I grew up in \textcolor{red}{Beijing}. & No, I'm an accountant in this company.  \\
\midrule
CLV(N=4) & I like \textcolor{red}{music} very much and I like dancing. How about you? & Yes, I'm from \textcolor{red}{Beijing}. I'm a \textcolor{red}{primary school teacher}. & Yes, I am a special policeman, and often travel, because my work is very tiring, I am suitable for \textcolor{red}{legal affairs}.  \\
 \bottomrule
 \end{tabular}
 \caption{A case study. Keywords are marked in red.}
\label{tab:case study}
\end{table*}

\section{Further Analysis}\label{sec:fa}

We further describe our model through a series of analyses. All analyses are based on the ConvAI2 dataset, and similar phenomena can be observed on Baidu PersonaChat.

\begin{table}[t!]
\centering
\small
\resizebox{\columnwidth}{!}{
\begin{tabular}{lrrr}
\toprule
Models & BLEU-1 & C-Dist-1/2 & Coh-Con.Score \\ \midrule
MSP (N=4) & \textbf{11.85} &  \textbf{5.63/26.91} & \textbf{23.01} \\
\quad\textit{w/o} Self-Separation & 6.74 &  7.31/28.02 & 13.17 \\
\quad\textit{w/o} Contrastive Learning & 9.36 & 7.13/27.92 & 16.17 \\
\quad\textit{w/o} Decider & 10.01 &  4.99/24.89 & 17.59 \\
\quad\textit{w/o} Joint Training & 9.69 & 5.09/24.71  & 18.16 \\ 
\bottomrule
\end{tabular}}
\caption{Ablation experiments results on ConvAI2.}
\label{tab:as}
\end{table}

\paragraph{Ablation Study} \label{sec:as}
To investigate the effects of different modules in CLV, we conducted an ablation study by removing modules. The results of the ablation study are shown in Table~\ref{tab:as}. We \textbf{first} investigated the impact of the core mechanism of the model, the self-separation algorithm. After removing the complete self-separation mechanism, the model degenerates to the most basic GPT-2 model, and it can be observed that the performance is on par with GPT-2. If we just remove the contrastive learning in the self-separation algorithm and keep the CVAE, we can see that the performance of the model also has a large decline, but the model's C-Dist-1/2 has an improvement, which is due to the global diversity due to the randomness of the sampled hidden variables in CVAE, which also indicates that CLV does sacrifice global diversity for other performance. \textbf{Then}, for the \textit{decider}, we eliminate persona information by directly computing the mean of the grouped persona information latent variables, and we can find that the \textit{decider} also plays an important role in CLV, especially when many dialogues are generated without considering persona, which shows that our decider can make decisions automatically. \textbf{Finally}, we conducted an experiment to validate our proposed joint training, and its performance degradation shows that it is difficult for the decider to learn how to make decisions without additional supervised signals. 

\begin{figure}[t!]
    \centering
    \includegraphics[width=\linewidth]{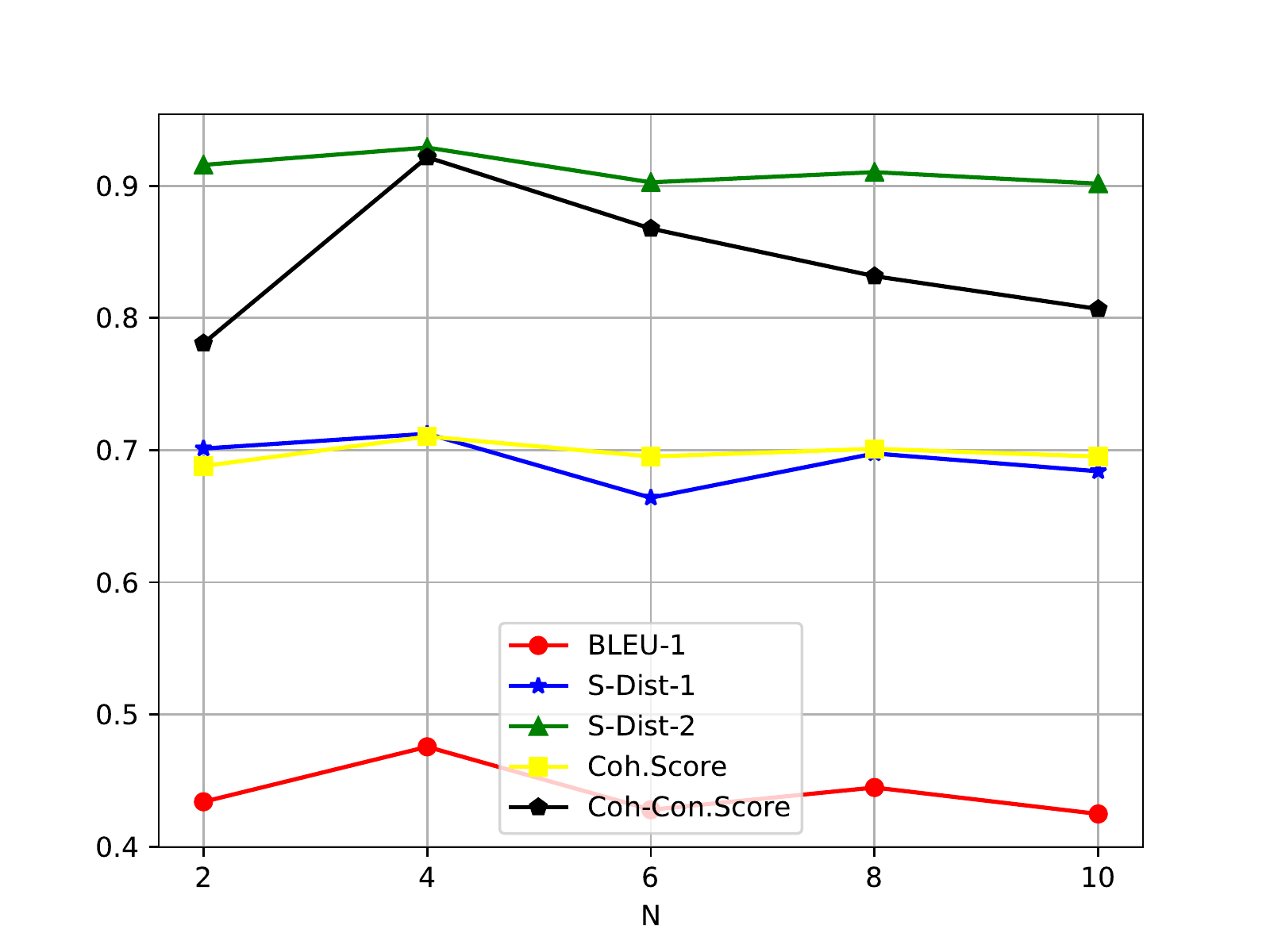}
    \caption[Caption for LOF]{Experiments with the different N on the ConvAI2 dataset. For ease of viewing, BLEU-1 and Coh-Con.Score are multiplied by a factor of 4.}
    \label{fig:n}    
\end{figure}

\paragraph{Effect of Self-Separation Coefficients} 
In CLV, the self-separation mechanism categorizes the persona information in an approximate implicit clustering way, and the self-separation coefficient N corresponds to the number of categories in the clusters. Intuitively, the self-separation factor will affect the model's performance, and we report this effect in Figure~\ref{fig:n}. 
The self-separation mechanism cannot do much good when the N is small. When N is set too large, the decider is also unable to make good decisions, which is due to the increased noise caused by too many categories, making the persona information too scattered, which is also consistent with the fact that the descriptive texts are always confined to several fixed perspectives.


To demonstrate the model's effectiveness more concretely, we conduct case studies. The results are shown in Table ~\ref{tab:case study}, which show that CLV can extract personal information, reconstruct persona profiles from queries alone, extract personal information, and generate fluent, personalized responses.

In Case 1, both CLV and BoB accurately answered "music" when asked about their hobbies, while CLV also used "How about you? " to keep the conversation going. In Case 2, CLV not only answered the address accurately but also flexibly used "school teacher" and "Affiliated Primary School of Renmin University" in the persona information to generate the response. In Case 3, all four models failed to accurately answer the question consistent with personality, but CLV still connected "lawyer" and "legal affairs".

By observing Cases 1 and 2, we can see that CLV can balance consistency and coherence, and its generation is consistent with persona and maintains context coherence. GPT-2 can only achieve basic sentence fluency. BoB and MSP can also generate good answers due to the help of context in reasoning. In Case 3, CLV creates a slightly fit answer, which is also better than the other models.

\section{Conclusion}
In this work, we propose a CLV model for personalized dialogue generation. Unlike existing works, we integrate the advantages of sparse and dense persona information. We use a \textit{self-separation} mechanism to implicitly cluster the persona information in the dense persona information text so that the \textit{decider} can consider different sparse categories of persona information during dialogue and enhance the personalization of dialogue generation. We also propose a more effective evaluation metric framework for personalized dialogue generation. The experimental results confirm the effectiveness of the model in generating personalized responses.

\section*{Limitations}
First, our model is a method of approximating clustering by contrastive learning, but due to the limitations of the model structure, we cannot directly explore the performance of past clustering algorithms on this task. Secondly, due to the large scale of the experiment, our dialogue generator only considers GPT-2. Although the ablation study proves the effectiveness of our model, it is a limitation. Finally, this paper proposes a complete evaluation framework for personalized dialogue generation. It is very effective, but the specific indicators in it still need to be discussed and further studied. In addition, the model assumes that response and persona are independent Gaussian distributions in CVAE. Although it performs well in the experiment, it does not conform to realistic cognition.

\section*{Ethics Statement}
From a general moral point of view, the generation of personalized dialogue in a broad sense may indeed cause problems such as identity forgery and the spread of false information. However, in this study, personalized corpus and responses are limited to the scope of experiments, which are not enough to threaten the real conversation. 

Furthermore, all models in this paper are trained on public corpus. The used datasets do not contain unethical language. We also ensure the anonymization of the human evaluation. 

\section*{Acknowledgements}
This work was supported by National Natural Science Foundation of China(62272340, 61876128, 61876129, 62276187, 61976154, 61402323), State Key Laboratory of Communication Content Cognition(Grant No.A32003).

\bibliography{anthology,custom}
\bibliographystyle{acl_natbib}

\appendix

\section{Appendix}
\label{sec:appendix}

\subsection{Default Parameter Settings}\label{sec:dps}
Our experiments are done based on pre-trained GPT-2, and we tried various model structures and hyperparameters, and the final hyperparameters are as follows: the size of GPT-2 embedding and GPT-2 hidden vector is 768. All word embedding dimensions are set to 768, and we use word2vec to initialize word embedding. The number of layers of Transformer is 12. The self-separation coefficient N is set from 2 to 16(default is 4), the MLP input dimension and output dimension in the model are kept the same as the hidden vector, and the number of batches was set to 16. The maximum learning rate is 1e-4. The training of the proposed model was done on an Nvidia Telsa V100 16G GPU. The total training time takes approximately 10 hours. The temperature hyperparameter $\tau$ is $0.5$. The pre-trained models used in these experiments of this paper include gpt2\footnote{\url{https://huggingface.co/gpt2}}, gpt2-chinese-cluecorpussmall\footnote{\url{https://huggingface.co/uer/gpt2-chinese-cluecorpussmall}}, xlm-roberta-base\footnote{\url{https://huggingface.co/xlm-roberta-base}}, and chinese-roberta-wwm-ext\footnote{\url{https://huggingface.co/hfl/chinese-roberta-wwm-ext}}.

We use kernel sampling~\citep{Holtzman2020The} as our decoding strategy, use the Adam~\citep{Kingma-2014-adam} optimizer to train the model and use AdamW~\citep{loshchilov2018decoupled} to warm up the generator. Please refer to the published project for additional details, which is publicly available\footnote{\url{https://github.com/Toyhom/CLV}}.


\subsection{NLI Model}\label{sec:nli}
NLI model is a triple classification model and can be design as: 
\begin{equation}
\begin{aligned}
&\text{NLI}(P,Q,R) \\ &=\left\{
\begin{aligned}
2  , &\text{if}\ P\ is\ consistent\ with\ R\ \\ &and\ Q\ is\ coherent\ with\ R\ , \\
1  , &\text{if}\ P\ is\ consistent\ with\ R\ \\ &but\ Q\ is\ not\ coherent\ with\ R\,   \\
0  , &\text{otherwise},
\end{aligned}
\right.
\end{aligned}
\end{equation}
Here NLI~\citep{welleck-2019-dialogue} is a pre-trained RoBERTa model~\citep{liu-2019-roberta}, fine-tuned using a dataset constructed based on ConvAI2 and Baidu PersonaChat, and the test set accuracy of NLI model on Chinese and English is 83.2\% and 83.1\%, respectively.

\end{document}